\PassOptionsToPackage{unicode}{hyperref}
\PassOptionsToPackage{hyphens}{url}
\documentclass[12 pt,]{article}
\usepackage{lmodern}
\usepackage{amssymb,amsmath}
\usepackage{ifxetex,ifluatex}
\ifnum 0\ifxetex 1\fi\ifluatex 1\fi=0 
  \usepackage[T1]{fontenc}
  \usepackage[utf8]{inputenc}
  \usepackage{textcomp} 
\else 
  \usepackage{unicode-math}
  \defaultfontfeatures{Scale=MatchLowercase}
  \defaultfontfeatures[\rmfamily]{Ligatures=TeX,Scale=1}
\fi
\IfFileExists{upquote.sty}{\usepackage{upquote}}{}
\IfFileExists{microtype.sty}{
  \usepackage[]{microtype}
  \UseMicrotypeSet[protrusion]{basicmath} 
}{}
\makeatletter
\@ifundefined{KOMAClassName}{
  \IfFileExists{parskip.sty}{%
    \usepackage{parskip}
  }{
    \setlength{\parindent}{0pt}
    \setlength{\parskip}{6pt plus 2pt minus 1pt}}
}{
  \KOMAoptions{parskip=half}}
\makeatother
\usepackage{xcolor}
\IfFileExists{xurl.sty}{\usepackage{xurl}}{} 
\IfFileExists{bookmark.sty}{\usepackage{bookmark}}{\usepackage{hyperref}}
\hypersetup{
  pdftitle={Incomplete Funnel Metrics Imputation in Large-Scale Online Controlled Experimentation},
  pdfauthor={Sumin Shen, Huiying Mao, Zezhong Zhang, Zili Chen, Keyu Nie, and Xinwei Deng; },
  hidelinks,
  pdfcreator={LaTeX via pandoc}}
\usepackage[margin=1in]{geometry}
\usepackage{graphicx,grffile}
\makeatletter
\def\maxwidth{\ifdim\Gin@nat@width>\linewidth\linewidth\else\Gin@nat@width\fi}
\def\maxheight{\ifdim\Gin@nat@height>\textheight\textheight\else\Gin@nat@height\fi}
\makeatother
\setkeys{Gin}{width=\maxwidth,height=\maxheight,keepaspectratio}
\makeatletter
\def\fps@figure{htbp}
\makeatother
\usepackage{bm}
\usepackage{dsfont}
\usepackage{pdfpages}
\usepackage{amsmath,amssymb}
\usepackage{empheq}
\usepackage{booktabs}
\usepackage{longtable}
\usepackage{array}
\usepackage{adjustbox}
\usepackage{caption}
\usepackage{algorithm}
\usepackage{algcompatible}
\usepackage{mathtools}
\usepackage{xcolor}
\usepackage{algorithm}
\usepackage{algcompatible}
\usepackage{booktabs}
\usepackage{longtable}
\usepackage{array}
\usepackage{multirow}
\usepackage{wrapfig}
\usepackage{float}
\usepackage{colortbl}
\usepackage{pdflscape}
\usepackage{tabu}
\usepackage{threeparttable}
\usepackage{threeparttablex}
\usepackage[normalem]{ulem}
\usepackage{makecell}
\usepackage{xcolor}
\usepackage[round]{natbib}  

\usepackage{comment}
\usepackage{algorithm}
\usepackage{algcompatible}
\usepackage{bbold}
\usepackage{bm}
\AtBeginDocument{%
  \providecommand\BibTeX{{%
    \normalfont B\kern-0.5em{\scshape i\kern-0.25em b}\kern-0.8em\TeX}}}

\title{Clustering-based Imputation for Dropout Buyers in Large-scale Online Experimentation}
\author{Sumin Shen\footnote{eBay China Center of Excellence, Shanghai 201203 (E-mail: sumshen@ebay.com)} \hspace{1cm}Huiying Mao\footnote{eBay Inc., San Jose, CA 95125 (E-mail: humao@ebay.com)}\hspace{1cm} Zezhong Zhang\footnote{eBay Inc., San Jose, CA 95125 (E-mail: zezzhang@ebay.com)} \\ Zili Chen\footnote{eBay China Center of Excellence, Shanghai 201203 (E-mail: zilchen@ebay.com)}\hspace{1.5cm} Keyu Nie\footnote{eBay Inc., San Jose, CA 95125 (E-mail: keyunie@gmail.com)}\hspace{1.5cm} Xinwei Deng\footnote{Professor of Statistics, Co-Director of Statistics and Artificial Intelligence Laboratory, Virginia Tech, Blacksburg, VA 24061 (E-mail: xdeng@vt.edu)}}
\date{Third Version, March 2023}

\begin{document}
\maketitle


\begin{abstract}
In online experimentation, appropriate metrics (e.g., purchase) provide strong evidence to support hypotheses and enhance the decision-making process.
However, incomplete metrics are frequently occurred in the online experimentation, making the available data to be much fewer than the planned online experiments (e.g., A/B testing). 
In this work, we introduce the concept of dropout buyers and categorize users with incomplete metric values into two groups: visitors and dropout buyers.
For the analysis of incomplete metrics, we propose a clustering-based imputation method using $k$-nearest neighbors.
Our proposed imputation method considers both the experiment-specific features and users' activities along their shopping paths, allowing different imputation values for different users. 
To facilitate efficient imputation of large-scale data sets in online experimentation, the proposed method uses a combination of stratification and clustering.
The performance of the proposed method is compared to several conventional methods in both simulation studies and a real online experiment at eBay.
\end{abstract}

{\bf KEY WORDS:} Experimentation; Metrics; Imputation; Clustering; A/B testing. 

\maketitle

\section{Introduction}
Online experimentation has been playing a key
role in data-driven decision making in the IT industry including Microsoft \citep{kohavi2009online, kohavi2014seven}, Google\citep{tang2010overlapping}, Linkedin \citep{xu2018sqr}, Netflix \citep{xie2016improving}, Uber, eBay \citep{nie2020dealing}, and many others \citep{gupta2019top}. 
Generally, online controlled experimentation, also known as A/B testing, is conducted for a
pre-determined amount of time to compare the difference in metrics between the
treatment group and the control group where users are randomly assigned
to. Prior to experimentation, a set of high-quality metrics are determined 
to assess the effects of new features in the treatment group. 
The collected metric results can provide strong evidence to support hypotheses and hence accelerate the decision-making process \citep{deng2016data, machmouchi2016principles, dmitriev2016measuring}.
However, incomplete metrics are frequently occurred in the online experimentation, making the available data to be much fewer than the planned A/B testing. 
In this work, our focus is on the analysis of metrics that have incomplete measurements at the end of data collection in experiments.

According to the positions in the shopping funnel, metrics can be categorized as top, middle, and bottom funnel metrics. 
%
For instance, a successful purchase typically requires users to
take multiple steps from the top homepage webpage to the bottom purchase webpage in the shopping funnel.
In online experimentation, it is common for millions of users to arrive at
the top funnel (e.g., homepage webpage), while only a small percentage of users reach the bottom funnel (e.g., purchase webpage). Between the transition from the top funnel to the
bottom funnel, users need to navigate through multiple pages where
they can exit from the shopping process. There are numerous scenarios in which
users can exit the funnel, resulting in incomplete records of their purchases or
other metrics. 
A common occurrence is simply that each experiment has its own experiment duration. Keeping experiments alive for a long period of time is expensive due to the high operational efforts and business opportunity costs. When we close experiments, we stop the track of all users, but some users might yet complete their purchases.
This incompleteness in metrics due to the delay in collecting
measurements for bottom-funnel metrics in experimentation are
inevitable. 
There is also the possibility that users are lost to follow due to technical issues or user unavailability. 
For instance, when users switch from the desktop app to the mobile app, they become unavailable. 
It is essential to fill in the incomplete metrics to improve
metric quality, leading to trustworthy results and better decisions.

With incomplete metric measurements, the inference of the difference
in metrics between the treatment and the control in experiments is at
the risk of being inaccurate \citep{imbens2001estimating, imbens2000analysis,goldstein2007subtle}. 
To analyze experiments with missing metric values, a naive approach is to disregard users with incomplete outcomes. This approach assumes that the
missingness is completely at random and that the fully observed users are
representative of the entire population. Such an approach will reduce the total number of users in the study, leading to a
decrease in the experiment power. The power decrease is substantial especially when the proportion of missingness is high. 

Various imputation methods have been developed to address problems with missing data.
One widely used method is the single imputation method, which fills in missing values with a single value, such as the mean of observed outcomes, for both the treatment group and the control group. 
The single imputation method preserves the full sample size, but it raises concerns regarding results with
a distorted distribution and underestimated uncertainty \citep{spineli2020comparison}. 
In addition, the single imputation method disregards information from other observed variables collected along users' journeys within the funnel. 
Other imputation methods have been developed for missing at random
(MAR) and missing not at random (MNAR) scenarios. The MAR assumes that the missing mechanism is only associated with the observed
variables \citep{rubin1976inference, imai2009statistical, bhaskaran2014difference}. 
Likelihood-based methods, such as generalized linear mixed models, are developed in clinical trials with incomplete outcomes \citep{molenberghs2004analyzing}. 
The performance of the methods depends
on the degree to which the assumptions are held for MAR. 
For MNAR in which the effect from missing outcomes is non-ignorable, the observed difference would be a biased estimate of the average treatment effect \citep{molenberghs2004analyzing}.
Regression-based imputation
methods, such as the logistic regression, are employed for modeling the
indicator for missingness \citep{mao2021driving}. 
Other prevalent methods, such as
matching imputations, identify similar users from a set of variables.
In general, these imputation methods require the identification of users with missing outcomes and users with
outcomes as zero. 
In other words, general imputation methods are often not appropriate to handle certain online experimentation scenarios in which users' missing outcomes represent both missing cases and zero cases. 

To address the above challenges, 
we propose a clustering-based imputation method using $k$-nearest
neighbors (kNN) for the analysis of online controlled experimentation in the presence of incomplete metrics. 
The key idea of the proposed method is to identify and impute incomplete metrics
with users' neighbors by incorporating the structure information of data from online experimentation. 
Specifically, the proposed method consists of two steps. 
The first step is to partition the data set into clusters after the stratification of experiment-specific features, including the treatment assignment and the buyers' characteristics. 
In the second step, we perform the kNN-based imputation. 
Moreover, we divide users with missing outcomes into two categories: visitors and dropout buyers, such that the information of dropout buyers can be better utilized.
Note that our framework assumes that the treatment assignment and user covariates are
fully observed, whereas only the outcome at the bottom of the funnel has missing values. 
The proposed method has three key advantages. First, the proposed method uses the informative covariates during users' journeys in the
shopping funnel to impute incomplete metrics. Specifically, our method
evaluates the heterogeneous impact from different user segments on
missing rates in metrics. 
Second, the imputed values from our method are intuitive to understand.  
Lastly, our method employs stratification and clustering to alleviate the computation issues for large-scale data in online experimentation.

Throughout the paper, we consider the metric \textit{Purchase} as an example of the incomplete metric at the funnel's bottom for illustration. 
We also assume that the \textit{Purchase} is the only metric (i.e., outcome) of interest in the experiment. 
The rest of the paper is organized as follows. In Sections 2 and 3, we detail the problem formulation, the proposed method, and the estimation procedures.
In Section 4, we present simulations. A real case study is conducted in Section 5. We conclude this work with some discussion in Section 6.

\section{Problem Formulation}
In the context of online controlled experiments, we can classify users into three types based on their purchase behaviors: visitors, real buyers, and
dropout buyers. 
Visitors participate in experiments but do not make
contributions (e.g., purchases). 
Real buyers not only participate in experiments but also make
contributions (e.g., purchases). 
Dropout buyers could have made their contributions (e.g., completed their transactions) within the experimentation period but failed due to various reasons. 
For example, users could drop out of the experiment because of unexpected external payment issues. 
Another example is that the experiment lost users due to various technical issues.

Suppose there are $n$ users in an experiment, and let $y_i \in \{0,1\}$ denote whether the $i$-th user is a buyer or not.
That is, 
\begin{equation*}
y_i =
\begin{cases}
1, & \textrm{user } i \textrm{ is either a real buyer or a dropout buyer},\\
0, & \textrm{user } i \textrm{ is a visitor},\\
\end{cases}
\end{equation*}
and 
$y_i = \mathbb{1}(z_i > 0),$ where $z_i \geq 0$ denotes the purchase metric value of the $i$-th user, and $\mathbb{1}(\cdot)$ is the indicator function. 
We know for sure that user $i$ is a real buyer and the corresponding value amount if he/she has completed transaction(s) during the experimentation period. In other cases, it is ambiguous whether he/she is a dropout buyer or merely a visitor. Therefore, we use $y_i^{obs}$ and $z_i^{obs}$ if the $i-$th user is a real buyer and $y_i^{mis}$ and $z_i^{mis}$ to represent the ambiguous situation (i.e., could be a dropout buyer or a visitor). To clarify, 
\begin{equation*}
y_i =
\begin{cases}
y_i^{obs} = 1,  \textrm{user } i \textrm{ is a real buyer},\\
y_i^{mis} = 
\begin{cases}
1, \textrm{user } i \textrm{ is a dropout buyer},\\
0,  \textrm{user } i \textrm{ is a visitor}.\\
\end{cases}
\end{cases}
\end{equation*}

However, some practitioners arbitrarily treat all $y_i^{mis}$ and $z_i^{mis}$ as 0 without the diligence to distinguish between dropout buyers and visitors. Here, we denote such an arbitrary but simplified buyer indicator as 
\begin{equation*}
\tilde y_i =
\begin{cases}
1,  &  \textrm{user } i \textrm{ is a real buyer},\\
0,  &  \textrm{otherwise. }
\end{cases}
\end{equation*}
Their corresponding vectors are denoted as $\bm y = (y_1, ...,  y_n), \bm z = (z_1, ...,  z_n), \tilde{\bm y} = (\tilde y_1, ..., \tilde y_n).$
Additionally, let \(\bm x_{i}\) denote the relevant features for user \(i\), \(\bm x_{i} = (x_{i1}, \ldots, x_{ip}) \in R^p\), \(p \ge 1\), and let
\(\bm X = (\bm x_{1}, \ldots, \bm x_{n})^T\),
Without loss of
generality, we assume that \(p\) features are continuous variables.

Suppose there are $m$ real buyers among the total $n$ users, and without loss of generality, let us assume the first $m$ users are real buyers. Denote $n$ users' purchase and transactional amount during the experimentation period using
vectors
$$\bm y = (\bm y^{obs}, \bm y^{mis}) =(y_1^{obs},\dots, y_{m}^{obs}, y_{m+1}^{mis}, \dots, y_n^{mis}),$$ 
$$\bm z = (\bm z^{obs}, \bm z^{mis}) =(z_1^{obs},\dots, z_{m}^{obs}, z_{m+1}^{mis}, \dots, z_n^{mis}).$$

The problem of interest is to impute missing values $\bm y^{mis}$ and $\bm z^{mis}$ in the context of online experimentation. 
Among users with missing values, visitors are mixed with dropout buyers. Therefore, our proposed method is to firstly identify the candidates of dropout buyers (i.e., identifying the candidates of 1s in $\bm y^{mis}$) with the help of a classification model and then impute the $\bm y^{mis}$ and $\bm z^{mis}$ using an efficient cluster-based nearest neighbors-based approach. 

\hypertarget{the-proposed-method}{%
\section{The Proposed Method}\label{the-proposed-method}}

The objective of the imputation problem is to impute missing values such that
they are close to the underlying true data. 
The missing value imputation problem can be formulated as 
\[
\underset{ {\bm \hat{y}}^{mis}}{\text{min}} ~l( \hat{\bm y}^{mis}, \bm y^{mis}),
\] where \(l(\hat{\bm y}^{mis}, \bm y^{mis})\) is a loss function to
quantify the difference between the imputed missing values
\(\hat{\bm y}^{mis}\) and the underlying true values \(\bm y^{mis}\).

Imputing missing values with non-parametric methods such as the nearest neighbors algorithm in large-scale data sets is challenging due to the large computation requirements for distances between pairs of data points.
To solve this challenge, we propose to incorporate the data clustering patterns into
the imputation. In other words, we partition users into \(c\)
clusters and then perform imputations within each cluster. Thus, the
cluster-based imputation problem is described as
\begin{align}
&\underset{{\bm \hat{y}^{mis}}}{\text{min}} ~\sum_{h=1}^{c}\sum_{C(i)=h}^{} l( \hat{ y}^{mis}_i, y^{mis}_i), 
\\ ~&s.t. \sum_{h=1}^{c}\sum_{C(i)=h}^{}||\bm{x}_i - \bm \mu_h||_2^2 \le g, ~ i \in I, \notag
\end{align}
where \(\bm x_{i}\) denotes the features for user \(i\), and \(C(i)=h\) represents the user \(i\) with missing value
\(y^{mis}_i\) belongs to cluster \(h\) with the centroid \(\bm \mu_h\),
the constant \(g\) controls the within-cluster distances, and
\(||\cdot||_2\) is the L\(_2\)-norm. The set of indices \(I\) is defined
as \(I = \{i:y_i \text{ is missing}\}\).
The features are selected based on experiment owners' domain knowledge.
After imputing \(y^{mis}_i\), we can estimate the corresponding \(z^{mis}_i\) as well.

Note that it is unknown whether a user with an incomplete metric is a visitor or a dropout buyer. 
The dropout buyers are mixed with visitors because both do not have their purchase information recorded. To
address the challenge, in Section 3.1, we apply the logistic regression model to identify a certain portion of visitors and narrow down the candidates
of dropout buyers. 
Section 3.2 will detail the proposed cluster-based imputation.
Notice that the data set in online controlled experiments often is very large such that the conventional clustering methods cannot be conducted efficiently.
To alleviate the computation issue, Section 3.3 will consider a stratification-based clustering and describe how to choose the number of clusters.

\hypertarget{visitors-imputation}{%
\subsection{Identifying Dropout Buyer Candidates}\label{visitors-imputation}}

The practitioners' simplified buyer indicator $\tilde{\bm y}$ reveals partial information in the true buyer indicator $\bm y$. 
Therefore, a classification model based on $(\bm X, \tilde{\bm y})$ provides us with the likelihood 
of purchases. Users with a high likelihood but missing purchase records can serve as the candidates for dropout buyers. Since $\tilde{\bm y}$ is used as a substitution of $\bm y$, we call  $\tilde{\bm y}$ pseudo-response. 

Specifically, we propose to apply the
logistic regression model for the buyer identification. 
Denote the conditional probability for user $i$ as
\(p(\bm x_i) = Pr(\tilde y_i = 1 |\bm x_i)\), that is,  \[
\begin{aligned}
\tilde y_{i}|\bm x_i = \left\{
          \begin{array}{ll}
            1, &  \mbox{w.p. } p(\bm x_i),\\
            0, & \mbox{w.p. } 1-p(\bm x_i).
          \end{array}
        \right.
\end{aligned}
\] We model the conditional probability \(p(\bm x_i)\) with the logistic
model \(\log ( p(\bm x_i) /(1-p(\bm x_i))) = \bm x_i^{T} \bm \beta\) with
\(\bm \beta\) = \((\beta_{1},\ldots,\beta_{p})^T\). Note that the
features used in the logistic regression model are believed to be
closely related to users' purchase behaviors. A threshold is needed in
the logistic model for classification. One widely used
threshold value is 0.5. Customers can choose the percentage of TN in the whole samples as the threshold according to their domain knowledge. 

Comparing the model prediction and pseudo-response, Table 1 summarizes four types of classification results: false positive
(FP), true negative (TN), false negative (FN), and true positive (TP)
from the classification model. The FP indicates that the users with
pseudo-response as 0 should have purchase information. We use this
inconsistency to figure out the candidates of dropout buyers. That is,
the FP cases can be either visitors or dropout buyers. The TN suggests
the agreement that these users do not have purchases recorded. Thus, we
treat all TN cases as visitors. The FN and the TP are users recorded
with purchase behaviors, and hence they are real buyers, not dropout
buyers or visitors.

\begin{table*}[t]
\caption{\label{tab:unnamed-chunk-2}Summary of four categories of results in the logistic regression model.}
\centering
\begin{tabular}[t]{l|r|r|l}
\toprule
 & Pseudo-response ($\tilde y$) & Prediction & Description\\
\midrule
True Negative (TN) & 0 & 0 & Visitors\\
\cline{1-4}
False Positive (FP) & 0 & 1 & Candidates of dropout buyers\\
\cline{1-4}
False Negative (FN) & 1 & 0 & Real buyers\\
\cline{1-4}
True Positive (TP) & 1 & 1 & Real buyers\\
\bottomrule
\end{tabular}
\end{table*}

Suppose there are $r$ visitors and $n-m-r$ dropout buyer candidates that have been identified. Without loss of generality, let us assume the first $r$ users in the missing set are those visitors. Then we write $\bm y^{mis}$ as 
\[\bm y^{mis} = (\bm y^{est}, \bm y^{*}) =(y_{m+1}^{est},\dots, y_{m+r}^{est}, y_{m+r+1}^{*}, \dots, y_n^{*}),\]
where \(y_i^{est} = 0, i = m+1, \ldots, m+r\), and \(y^{*}_i \in \{0, 1 \}, i = m+r+1,\ldots,n\)
with 0 representing visitors and 1 representing dropout buyers.
Similarly, we denote the corresponding continuous response for the purchase amount
as
\[\bm z^{mis} = (\bm z^{est}, \bm z^{*}) =(z_{m+1}^{est},\dots, z_{m+r}^{est}, z_{m+r+1}^{*}, \dots, z_n^{*}),\]
where \(z_i^{est} = 0, i = m+1, \ldots, m+r\), represents the purchase
amounts from estimated visitors and \(z_i^{*}\) represents the missing
non-negative response from \(n-m-r\) users. 
In the following imputation methods in Section \ref{cluster-based-imputation}, we consider
\(\bm z^{est}\) to be zeros and our major focus is to impute \(\bm z^{*}\).

\hypertarget{cluster-based-imputation}{%
\subsection{Clustering-based Imputation for Dropout Buyers}\label{cluster-based-imputation}}
To impute the missing purchase value \(\bm z^{*}\) of the dropout buyers, we adopt the clustering-based method using kNN techniques.
It is noted that clustering improves data analysis efficiency by identifying inherent structure patterns and partitioning the large-scale data set into small subsets.
In each strata \(\bm X_{tu}\) (described later in the stratification step), we perform the $K$-means clustering method \citep{macqueen1967some} to form clusters, which
is formulated as \[
\underset{C}{\text{minimize}}\sum_{h=1}^{c}\sum_{C(i)=h}^{}||\bm{x}_i - \bm \mu_h||_2^2.
\]

On top of clustering, we use the triangle inequality rule (described later) to ensure the consistent identification of nearest neighbors in the $k$-nearest neighbors (kNN) approach for imputation. The main idea of the kNN method is that nearby data points are similar to each other. The kNN algorithm is straightforward
and does not require parametric model estimation, but it is
computationally expensive and becomes slow as the size of the data set
increases. However, this computational burden is greatly mitigated by
the strategy of clustering. Given the specific cluster \(h\) (i.e., the fixed constraint in (1)), 
the imputation problem (1) with the kNN method can be written as
\begin{align}
y_i^* = \underset{L}{\text{argmax}} \sum_{x_j \in N_k(x_i)} \mathbb{1}\{y_j = L\}, ~i \in I ,~C(i)=h,
\end{align}
where \(L \in \{0,1\}\) is the binary label, \(k\) is a positive integer representing the size of target user's nearest neighbors \(N_k(x_i)\) and \(j\) is the nearest neighbors' user index. 
The performance of the kNN method may be affected by different $k$ values. The optimal $k$ value depends on the underlying structure of data sets.
In this work, we use a fixed value 15 for $k$. 
It is not difficult to derive the solution to the objective function, which is written as \[
\hat{y}^{*}_i = 
\begin{cases}
    1,& \sum_{j=1}^{k}y_j/k >= 0.5, \\
    0,              & \textrm{otherwise},
\end{cases}
\] 
where \(\sum_{j=1}^{k}y_j/k\) is the average of response \(y_j's\) in the nearest neighbors.

With the imputed \(\hat{y}^{*}_i\), we obtain the corresponding imputed missing value \(\hat{z}^{*}_i\) from the cost function formulated as 
\[
\begin{aligned}
\underset{{z}^{*}_i}{\text{minimize}} ~& \hat{y}_i^* \sum_{j=1}^{k} ||{z}^{*}_i - z_j||_2^2 + (1-\hat{y}_i^*)||{z}_i^{*}||_2^2,\\
~ & ~z_i^{*}\ge 0, ~i \in I,~C(i)=h, 
\end{aligned}
\]
That is, the estimated \(\hat{z}_i^{*}\) is given by \[
\hat{z}^{*} = 
\begin{cases}
    \sum_{j=1}^{k}z_j/k,& \sum_{j=1}^{k}y_j/k >= 0.5, \\
    0,              & \textrm{otherwise},
\end{cases}
\]
where \(\sum_{j=1}^{k}z_j/k\) is the average of response \(z_j's\) in the nearest neighbors.

The nearest neighbors are determined based on their distances
to the target user, that is, the \(k\) closest neighbors are found by \[
min~\sum_{j=1}^{k} d(\bm {x}_i, \bm x_j), 
\] where \(d(\bm x_i, \bm x_j)\) is the distance between the users \(i\)
and \(j\). 

To further accelerate the computation, we adopt the triangle inequality rule \citep{wang2011fast}, which avoids unnecessary distance calculations. 
We first obtain the \(k\) nearest neighbors within the closet cluster and denote their largest distance as \(d_{max}\). 
We denote the distance between the target user and any other cluster centroid as \(d_{1}\), the distance between any user in the same cluster and its cluster centroid as \(d_{2}\), the distance between the target user and any user as \(d_{3}\).
The idea of the triangle inequality rule is that when \(d_{max} \le\) |\(d_{1}\) - \(d_{2}\)|, then \(d_{max}\le\) \(d_{3}\). As a result, we do not have to explicitly calculate \(d_{3}\), which greatly speeds up the distance computation and ensures that the identification of nearest neighbors is robust to clustering. 
In this study, we use the L\(_2\)-norm to measure distances.

\hypertarget{stratification}{%
\subsection{Efficient Clustering Strategy}\label{stratification}}

Note that the data set in the online controlled experimentation often is very large to cluster in the imputation step. 
To reduce the computational burden in clustering, we propose the stratification-based clustering approach. The key idea is to firstly stratify the user pool, and then perform clustering within each strata. 

In the stratification step, we stratify users into two hierarchical
levels: treatment assignment and users' buying characteristics. The
treatment assignment, including the treatment group and the control
group, is determined by the experimentation configuration. Generally, in
online controlled experiments there are two treatment assignments:
control and treatment. However, more than two treatment assignments are
possible in cases such as multivariant experiments. User's buying
characteristics, including new buyers, infrequent buyers, frequent buyers, and idle buyers, are categorized based on users' purchase
activities at eBay. There are in total 12 buyer categories. Note that
both the experimentation configuration and the users' buying segments
are determined prior to the start of the experimentation. The
hierarchical stratification is formulated as \[
\bm X = \bigcup_{t=1}^{T}\bigcup_{u=1}^{U}\bm X_{tu},
\] where \(\bm X_{tu}\) is the strata at
the \(t\)-th treatment level and the \(u\)-th users' buying
characteristics in the feature space \(\bm X\), and there are in total \(T\) levels treatment assignment and \(U\)
levels users' buying characteristics. 

The combination of stratification and clustering within each strata
greatly improves computation efficiency in the imputation step, where
the neighbors of the target user are searched within all clusters.

The number of clusters in each strata from the stratification is
obtained by maximizing a simplified version of the Silhouette score, also known as simplified Silhouette. The Silhouette score is an effective
measure of clustering goodness \citep{rousseeuw1987silhouettes},
but it requires an intense computation of the distance betweeb each data point and the rest data points. The simplified Silhouette improves the computational efficiency of the Silhouette score by calculating the distances between each data point and centroids of clusters \citep{hruschka2004evolutionary}. The simplified Silhouette of data point \(i\), denoted as \(SS_i\), is defined as \[
\text{SS}_i = \frac{b_i-a_i}{max(a_i,b_i)},
\] where \(a_i\) is the distance between the data point \(i\) and the
centroid of the cluster it belongs to, and \(b_i\) is the minimum of
distances between the data point \(i\) and the centroids of other
clusters. The final simplified Silhouette is the average of all data
points' simplified Silhouette. Note that the distances of each data point to its
cluster centroid have already been calculated and recorded during the modeling process of k-means clustering, which greatly reduces the computational burden of the simplified Silhouette.

A pseudo-code for the proposed method is summarized in Algorithm 1.

\begin{algorithm}
  \caption{Pseudo code for the proposed method}
  \begin{algorithmic}[1]
  \linespread{1.0}\selectfont
    \STATE INPUT: the binary response $\bm y = (\bm y^{obs}, \bm y^{mis})$, the continuous response $\bm z = (\bm z^{obs}, \bm z^{mis})$, the pseudo-response $\tilde {\bm y}$, and the predictor features $\bm X$.
    \STATE Perform the logistic regression model on the data set with the pseudo-response $\tilde {\bm y}$ and the predictor features $\bm X$. Obtain the classification results including false positive (FP) and true negative (TN).
    \STATE \textbf{Stratification}. Stratify $\bm X$ based on the treatment assignment and the users' buying characteristics. 
    \FOR {each strata}
    \STATE Use the FP as the test set, and the rest as the training set in the kNN method.
    \FOR {each target user in the test set}
    \STATE \textbf{Clustering}. Perform k-means clustering in the strata, 
    \STATE Find out the cluster that the target user belongs to,
    \STATE \textbf{Imputation}. Within that cluster, find the initial $k$ nearest neighbors and its corresponding maximum distance $d_{max}$.
    \FOR {each user in other clusters from the farthest to the nearest to the cluster centroid}
    \STATE Obtain the distance between the target user and this cluster centroid \(d_{1}\), and the distance between this user in this cluster and its cluster centroid \(d_{2}\),
    \IF {\(d_{max} \le\) |\(d_{1}\) - \(d_{2}\)|}
    \STATE Move on to the next cluster.
    \ELSE 
    \STATE Update the $k$ nearest neighbors and $d_{max}$.
    \ENDIF
    \ENDFOR
    \IF {$\sum_{l=1}^{k}y_l/k >= 0.5$}
    \STATE Impute $\hat y^{*} = 1$ and $\hat z^{*} = \sum_{l=1}^{k}z_l/k$.
    \ELSE {}
    \STATE Impute $\hat y^{*} = 0$ and $\hat z^{*} = 0$.
    \ENDIF
    \ENDFOR
    \ENDFOR 
    \STATE OUTPUT: $\bm z = (\bm z^{obs}, \bm {\hat z}^{mis})$
  \end{algorithmic}
\end{algorithm}

\hypertarget{performance-evaluation}{%
\section{Simulation}\label{performance-evaluation}}
In this section, we conduct the simulation studies to evaluate the performance of the proposed cluster-based KNN imputation method. The complete response has two parts: the non-zero part and the zero part. The non-zero part of response follows a Gaussian distribution  \( z_s = 1.5 + 1.1 w + 1.1 x_{s1} + 0.2 x_{s2} + \epsilon\), $\epsilon \sim N(0, 0.25)$ where $w$ is the binary assignment to the control and the treatment group, and $x_{s1}$ and $x_{s2}$ are variables normally distributed $N$(0.1, 1) and $N$(0.2, 2.25), respectively. The binary indicator of response follows a Bernoulli distribution with the conditional probability \(Pr(y_s=1|x_{s3})\) expressed by a logistic regression model \(logit(x_{s3}) = -1 + 5.8 x_{s3}\), where $x_{s3}$ is a variable with a Gaussian distribution $N$(0.2, 0.04). 
In the simulation, we consider three scenarios for generating missing values in the response for both the control and the treatment group. 
In scenario 1 (S1), the missing is completely at random. 
In scenario 2 (S2), the missing probability is described with a logistic regression model depending on an unobserved variable following a Gaussian distribution. 
In scenario 3 (S3), the missing is dependent on the value of the response. Specifically, the missing response is indicated if its value exceeds a pre-defined threshold within the control and the treatment group. 
In all three scenarios, we further treat responses with zero as missing to represent real cases where users have incomplete records. 
The sample size is fixed at 5000. 

We compare the proposed method with six benchmark models, including (i) Complete-case analysis (BM\(_1\)), (ii) Unconditional control-mean imputation (BM\(_2\)), (iii) Unconditional treatment-mean imputation (BM\(_3\)), (iv) Unconditional zero imputation (BM\(_4\)), (v) Best-case analysis (BM\(_5\)), (vi) Worst-case analysis (BM\(_6\)). 

Complete-case analysis removes cases with missing values and uses only
cases with complete outcomes. Specifically, we discard \(\bm z^{mis}\) and
the sample size is reduced to \(m\), that is,
\[
\text{BM}_1:~\bm z = \bm z^{obs}.
\] The complete-case analysis is easy to implement but generates unnecessary
waste of information especially when the number of incomplete cases is
substantial.

Unconditional control-mean imputation uses the mean in the observed
users in the control group to impute missing values while unconditional treatment-mean imputation uses the mean in the observed users in the
treatment group for imputation. That is,
\[
\begin{aligned}
\text{BM}_2: ~&\hat z^{*}_i = \frac{\sum_{c=1}^{n_c} z_c^{obs}}{n_c},~c \in C, \\
\text{BM}_3: ~&\hat z^{*}_i = \frac{\sum_{t=1}^{n_t} z_t^{obs}}{n_t},~t \in T,
\end{aligned}
\] where the set of indices \(C\) is defined as
\(C = \{c:z_c \text{ is in the control group.}\}\) and the set of
indices \(T\) is defined as
\(T = \{t:z_t \text{ is in the treatment group.}\}\). \(n_c\) and
\(n_t\) is the number of sample sizes in the control group and in the
treatment group, respectively. Unconditional zero imputation uses zero
to impute missing values, that is, \[
\begin{aligned}
\text{BM}_4: ~&\hat z^{*}_i = 0, ~i = m+r+1, \ldots, n. 
\end{aligned}
\] These three imputation methods are different types of single value
imputation approach, which can keep the full data size. But these
imputation methods treat the missing values as fixed, distorting the
distribution and ignoring the uncertainty in the missing values.

The best-case analysis imputes missing values in the treatment (control)
group with the mean in the users in the treatment (control) group. In
contrast to the best-case analysis, the worst-case analysis imputes missing values in the treatment (control) group with the mean in the users in
the control (treatment) group. Here, we assume that the testing feature
in nature has a positive impact, and thus the mean in the treatment group
is expected to be greater than the mean in the control group. 
As a result, the difference between the imputed missing values in the treatment group and the control group aligns with the feature impact in the best-case analysis, but contradicts the feature impact in the worst-case analysis.
The best-case analysis and the worst-case analysis are expressed as \[
\begin{aligned}
\text{BM}_5: ~&\hat z^{*}_t = \frac{\sum_{t=1}^{n_t} z_t^{obs}}{n_t},~  \hat z^{*}_c = \frac{\sum_{c=1}^{n_c} z_c^{obs}}{n_c},\\
\text{BM}_6: ~&\hat z^{*}_t = \frac{\sum_{c=1}^{n_c} z_c^{obs}}{n_c},~  \hat z^{*}_c =  \frac{\sum_{t=1}^{n_t} z_t^{obs}}{n_t},
\end{aligned}
\] where \(\hat z^{*}_t\) (\(\hat z^{*}_c\)) is the imputed missing
value in the treatment (control) group, \(z_t^{obs}\) (\(z_c^{obs}\)) is
the observed value in the treatment (control) group.

To check the performance of the proposed method, we estimate the mean and
variance in the control group, and compute lift in the mean between the treatment group
and the control group, the standard error (SE) of the difference between the treatment
and control group, coefficient of variation (CV) for the control group,
zero rate (ZR) and p-value. The lift in the mean between the treatment group
and the control group is described as \[
\text{Lift} = \frac{\mu_t - \mu_c}{\mu_c} \times 100\% = (\frac{\sum_{t=1}^{n_t} z_t^{}}{n_t} - \frac{\sum_{c=1}^{n_c} z_c^{}}{n_c})/\frac{\sum_{c=1}^{n_c} z_c^{}}{n_c} \times 100\%,
\] where \(\mu_t\) and \(\mu_c\) are the mean in the treatment group and
the control group, respectively.

The SE is expressed as \[
\text{SE} = \sqrt{\frac{(n_t-1)s^2_t + (n_c-1)s^2_c}{n_t+n_c-2}\cdot (\frac{1}{n_c} + \frac{1}{n_t})},
\] where \(s_t\) and \(s_c\) are the standard errors for the treatment
group and the control group, respectively.

In online experimentation, 
the faster we run experiments, the more economic
benefits, and less operational costs are achieved. Given constant user
traffic, running experiments faster means a smaller number of users
required \citep{wu2011experiments, deng2013improving}. The CV is
proportional to the number of users required for achieving a
pre-determined statistical power of experiments. The CV is expressed as \[
\text{CV} = \frac{s_c}{\mu_c}.
\] The smaller the CV, the smaller the user size required to detect the
difference at the specific statistical power, and thus the higher sensitivity.

The ZR is the ratio of the number of zero's (\(n_{zero}\)) in imputed
\(\bm z\) out of total data size \(n\), described as \[
\text{ZR} = \frac{n_{zero}}{n}.
\]
The ZR evaluates the proportion of visitors with the outcome as zero after the imputation method. 

We compare the performance of the proposed method and benchmark methods in all scenarios in Table 2. In S1 and S2, the proposed cluster-based KNN imputation method has the closest $\mu_c$, $\mu_t$ and ZR compared to the method NoMissing.
The BM\textsubscript{2} and BM\textsubscript{3} methods have larger $\mu_c$ and $\mu_t$ because these methods impute all missing values with nonzero values, which is indicated by their ZR values being 0. 
The proposed method has a comparable $s\textsubscript{c}$ value to the method NoMissing, while the BM\textsubscript{2}, BM\textsubscript{3}, BM\textsubscript{4}, and BM\textsubscript{5} methods have smaller values. This might be explained that the imputation values in the proposed method are not fixed as in the BM\textsubscript{2}, BM\textsubscript{3}, BM\textsubscript{4}, and BM\textsubscript{5} methods. 
Though the BM\textsubscript{1} method has a similar $s\textsubscript{c}$ compared to the proposed method, its sample size is smaller due to the removal of samples with missing responses.  
In S3, the proposed method does not outperform the BM\textsubscript{2} method. This is probably due to the fact that in S3 the missing response values can be partitioned into one particular group. When this entire group is missing, it is difficult for the KNN-based imputation approach to find good neighbors of missing responses. As a result, the estimated $\mu_c$ and $\mu_t$ are not close to the truth.   

{\footnotesize
\renewcommand{\arraystretch}{0.65}
\noindent\hspace{-0.04\textwidth}
\begin{minipage}[t]{1\textwidth}
\begin{longtable}{ccccccccc}
\caption{Performance comparisons of benchmark methods from 50 simulation replications (mean and standard errors (in parenthesis)). Note that method NoMissing uses the original complete response prior to the missing assignment.} \label{tab:unnamed-chunk-4}
\\\toprule
Scenario & Method & Lift (\%) & $\mu\textsubscript{c}$ & $\mu\textsubscript{t}$ & $s\textsubscript{c}$ & CV & $n\textsubscript{c}$ & ZR\\
\midrule
S1 & BM\textsubscript{1} & 65.6 (4.96) & 1.7 (0.05) & 2.8 (0.04) & \textbf{1.2} (0.03) & 0.7 (0.03) & 953.8 (30.33) & 0 (0)\\
 & BM\textsubscript{2} & 24.9 (2.24) & 1.7 (0.05) & 2.1 (0.03) & 0.8 (0.02) & 0.5 (0.02) & 2504.1 (28.23) & 0 (0)\\
 & BM\textsubscript{3} & 17.8 (1.14) & 2.4 (0.03) & 2.8 (0.04) & 0.9 (0.02) & 0.4 (0.01) & 2504.1 (28.23) & 0 (0)\\
 & BM\textsubscript{4} & 65.4 (9.85) & 0.6 (0.02) & 1.1 (0.04) & 1.1 (0.02) & 1.8 (0.04) & 2504.1 (28.23) & 0.6 (0.01)\\
 & BM\textsubscript{5} & 65.6 (4.96) & 1.7 (0.05) & 2.8 (0.04) & 0.8 (0.02) & 0.5 (0.02) & 2504.1 (28.23) & 0 (0)\\
& BM\textsubscript{6} & -11.1 (0.76) & 2.4 (0.03) & 2.1 (0.03) & 0.9 (0.02) & 0.4 (0.01) & 2504.1 (28.23) & 0 (0)\\
& Proposed & 40.3 (11.30) & \textbf{1.1} (0.25) & \textbf{1.5} (0.24) & 1.3 (0.20) & 1.2 (0.09) & 2504.1 (28.23) & 0.4 (0.01)\\
& NoMissing & 64.8 (7.41) & 0.9 (0.03) & 1.5 (0.04) & 1.2 (0.02) & 1.4 (0.03) & 2504.1 (28.23) & 0.5 (0.01)
\\
\midrule
S2 & BM\textsubscript{1} & 65.0 (4.41) & 1.7 (0.04) & 2.8 (0.04) & \textbf{1.2} (0.03) & 0.7 (0.03) & 958.6 (29.9) & 0 (0)\\

 & BM\textsubscript{2} & 24.8 (2.02) & 1.7 (0.04) & 2.1 (0.03) & 0.8 (0.02) & 0.5 (0.02) & 2504.1 (28.23) & 0 (0)\\

 & BM\textsubscript{3} & 17.8 (1.06) & 2.4 (0.03) & 2.8 (0.04) & 0.9 (0.03) & 0.4 (0.01) & 2504.1 (28.23) & 0 (0)\\

 & BM\textsubscript{4} & 64.3 (9.47) & 0.6 (0.02) & 1.1 (0.04) & 1.1 (0.02) & 1.7 (0.04) & 2504.1 (28.23) & 0.6 (0.01)\\

 & BM\textsubscript{5} & 65.0 (4.41) & 1.7 (0.04) & 2.8 (0.04) & 0.8 (0.02) & 0.5 (0.02) & 2504.1 (28.23) & 0 (0)\\

 & BM\textsubscript{6} & -11.0 (0.59) & 2.4 (0.03) & 2.1 (0.03) & 0.9 (0.03) & 0.4 (0.01) & 2504.1 (28.23) & 0 (0)\\
 & Proposed & 39.4 (10.79) & \textbf{1.1} (0.25) & \textbf{1.5} (0.24) & 1.3 (0.20) & 1.2 (0.09) & 2504.1 (28.23) & 0.4 (0.01)\\
 & NoMissing & 64.8 (7.41) & 0.9 (0.03) & 1.5 (0.04) & 1.2 (0.02) & 1.4 (0.03) & 2504.1 (28.23) & 0.5 (0.01)
\\
\midrule
S3 & BM\textsubscript{1} & 100.9 (8.84) & \textbf{1.1} (0.05) & 2.2 (0.03) & \textbf{0.9} (0.02) & 0.8 (0.05) & 958.6 (29.9) & 0 (0)\\

 & BM\textsubscript{2} & 38.4 (4.02) & \textbf{1.1} (0.05) & \textbf{1.5} (0.03) & 0.6 (0.02) & 0.5 (0.03) & 2504.1 (28.23) & 0 (0)\\

 & BM\textsubscript{3} & 23.7 (1.33) & 1.8 (0.03) & 2.2 (0.03) & 0.8 (0.02) & 0.4 (0.02) & 2504.1 (28.23) & 0 (0)\\

 & BM\textsubscript{4} & 100.1 (15.36) & 0.4 (0.02) & 0.8 (0.03) & 0.8 (0.02) & 1.8 (0.06) & 2504.1 (28.23) & 0.6 (0.01)\\

 & BM\textsubscript{5} & 100.9 (8.84) & \textbf{1.1} (0.05) & 2.2 (0.03) & 0.6 (0.02) & 0.5 (0.03) & 2504.1 (28.23) & 0 (0)\\

 & BM\textsubscript{6} & -14.7 (0.89) & 1.8 (0.03) & \textbf{1.5} (0.03) & 0.8 (0.02) & 0.4 (0.02) & 2504.1 (28.23) & 0 (0)\\
  & Proposed & 71.6 (9.67) & 0.6 (0.03) & 1.0 (0.03) & 0.8 (0.02) & 1.4 (0.05) & 2504.1 (28.23) & 0.5 (0.01)\\
 & NoMissing & 64.8 (7.41) & 0.9 (0.03) & 1.5 (0.04) & 1.2 (0.02) & 1.4 (0.03) & 2504.1 (28.23) & 0.5 (0.01)
\\
\bottomrule
\end{longtable}
\end{minipage}
}

\hypertarget{case-study-ranking-search-experiment}{%
\section{Case Study: Search Ranking Experiment
}\label{case-study-ranking-search-experiment}}

To illustrate the proposed method, this section uses a real online experiment whose objective was to improve eBay's
item ranking search results based on one ranking algorithm. 
The experiment hypothesis is that integrating information about negative buyer
experiences into the ranking algorithm will
reduce the visibility of items with a high probability of negative buyer
experiences in search
results, resulting in lower product return rates and increased revenues. This
experiment lasts three weeks. A portion of eligible eBay users are selected and randomized
into three variants -- two treatment groups and
one control group. The number of
participant users in each variant exceeds 10 million.
One of the most important outcomes is related to purchases, denoted here as PR.

The outcome PR is incomplete due to its high missing rate. The PR is recorded when users
made purchases during the experiment's data collection period, but not when either of the following occurred: users did not make purchases, or the platform was unable to record the purchases
before the end of the experiment's data collection period. To impute PR and thus identify visitors
and dropout buyers, we use these informative covariates, including the
treatment assignment, the number of sessions, the number of sessions
with searches, the number of sessions with qualified events highly related to purchases at eBay, and the user's buying characteristics.
The treatment assignment is pre-determined before
running the experiment to assign users to the treatment group and the
control group. The number of sessions corresponds to the number of sessions users
have throughout the experiment. The number of sessions with searches is the
number of sessions that contain at least one search activity. The
number of sessions with qualified events is the number of sessions that include at
least one qualified event activity. The buying characteristics of users are their
historical purchasing patterns at eBay. These useful covariates are
complete and do not have missing values. We impute the outcome PR using the proposed
cluster-based imputation method. In the step
of stratification, we divided the large-scale data set into
smaller subsets based on two variables: the treatment assignment and
user's buying characteristic. When performing clustering within each strata, we use the number of sessions, the number of sessions with searches, and the number of sessions with qualified events.

In Table 3, we compare the performance between the proposed cluster-based
imputation method and benchmark methods. The proposed method has a smaller mean in the control group and ZR than other methods except for the BM\(_4\).
The proposed imputation method identifies visitors and dropout buyers
from missing values. That being said, the proposed cluster-based
imputation method imputes zeros for visitors, which is a portion of
users with missing outcomes, and positive values for dropout buyers. Compared to the BM\(_4\), the proposed
imputation method has a smaller size of zero and thus a larger mean in the control group. Compared to other mean-imputation methods that
impute all missing values with a single value, the proposed
imputation method has more zero's and a smaller mean in the control group.
The proposed method has a larger CV in the control group than all other
methods, with the exception of BM\(_4\). This is largely attributable to the change in the
mean of the control group, as the pooled standard errors for all methods, with the exception of BM\(_4\), are quite close. The proposed method has the
smallest lift, and all methods have a consistent direction of lift. Based
on the p-value and the Type I error as 10\%, the proposed method and
BM\(_5\) are statistically significant, indicating that there is sufficient evidence to
reject the null hypothesis, whereas other methods are not statistically
significant. This is expected because it is well known that single
imputation methods tend to dilute mean differences, producing results that there is no difference
between the control group and the treatment group. The proposed method
has a larger variance in the control group and SE than other methods
except for the BM\(_1\). The BM\(_1\) has a reduced sample size, resulting in the largest variance and SE for the control group. Unlike other methods, with the exception of the BM\(_1\), the proposed method does not
ignore variance among missing values, resulting in a greater variance.
\begin{table*}[t]
\caption{\label{tab:unnamed-chunk-4}Performance comparisons of benchmark methods in the ranking search experiment. Note that the values of $s^2\textsubscript{c}$, $\mu\textsubscript{c}$, CV, and SE are not real and masked with particular linear transformation to meet the disclosure requirement.}
\centering
\begin{tabular}[t]{l|r|r|r|r|r|r|r}
\hline
Method & $s^2\textsubscript{c}$ & $\mu\textsubscript{c}$ & CV & ZR & Lift (\%) & SE & p-value\\
\hline
BM\textsubscript{1} & 107035.21 & 1235.8 & 0.265 & 0.00 & -0.37 & 0.33 & 0.17\\
\hline
BM\textsubscript{2} & 20003.17 & 390.5 & 0.362 & 0.00 & -0.16 & 0.06 & 0.28\\
\hline
BM\textsubscript{3} & 20004.96 & 389.9 & 0.363 & 0.00 & -0.17 & 0.06 & 0.28\\
\hline
BM\textsubscript{4} & 20693.30 & 213.7 & 0.673 & 0.83 & -0.29 & 0.06 & 0.31\\
\hline
BM\textsubscript{5} & 20003.17 & 390.5 & 0.362 & 0.00 & -0.29 & 0.06 & 0.05\\
\hline
BM\textsubscript{6} & 20004.96 & 389.9 & 0.363 & 0.00 & -0.03 & 0.06 & 0.82\\
\hline
Proposed & 21194.12 & 246.6 & 0.590 & 0.80 & -0.50 & 0.06 & 0.05\\
\hline
\end{tabular}
\end{table*}

Figure 1 illustrates the increase in the mean of the control group across users' buying segments for the proposed cluster-based imputation method and the
zero-imputation method. Different user segments have different mean
values, with the top two being the frequent buyer levels II and III.
The proposed imputation method has larger mean values than the zero imputation method in nearly all user segments. The segments 
the frequent buyer levels II and III have considerably larger mean increases than the idle buyer levels. This suggests that
the dropout buyers are more likely to occur in the frequent buyer levels II and III, while in the segments such as idle buyer levels, users
with unrecorded outcomes are more likely to be visitors. This is
consistent with the findings in Figure 2 regarding the allocation of the zero rate across
user segments. Different user segments have varying degrees of zero rate. The zero rates for frequent buyer levels II and III are approximately 45\%, whereas the zero rates for idle buyer levels II and III are above 90\%. This is reasonable given that frequent buyer levels II and III are more likely to make purchases, resulting in low zero values for outcome PR. The high zero rate corresponds to the
low mean value in Figure 1.

\begin{figure}[]
\caption{Comparison of mu across user segments between the proposed imputation method and the zero imputation method for the treatment group. The tick values in the vertical axis are omitted for the restriction of disclosure.}
{\centering \includegraphics[width=\linewidth]{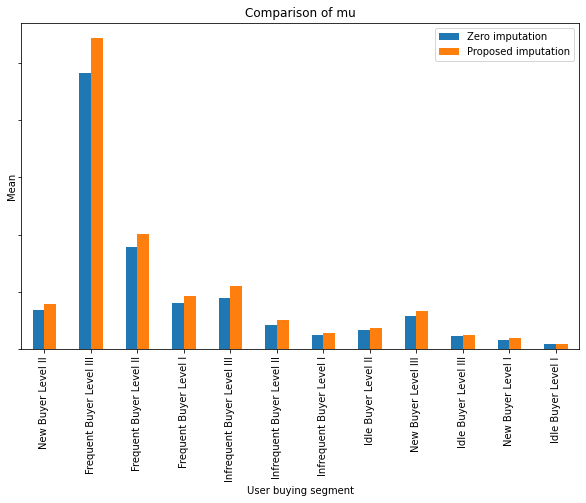}
}
\end{figure}

\begin{figure}[]
\caption{Comparison of zero rate across user segments between the proposed imputation method and the zero imputation method.}
{\centering \includegraphics[width=\linewidth]{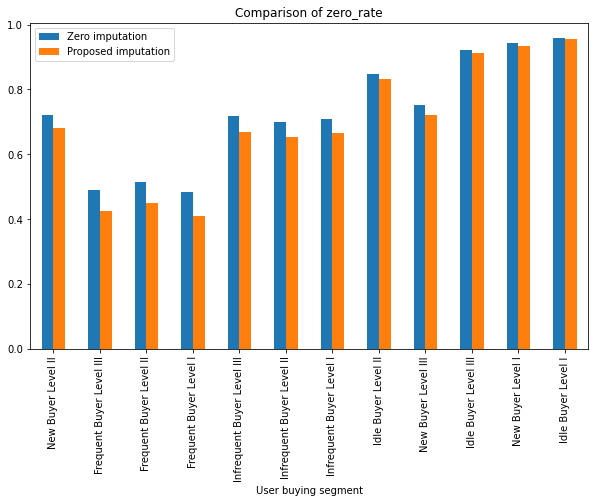}
}
\end{figure}

Figure 3 shows the distribution of CV across user segments for the
proposed imputation method and the zero imputation method. 
For both methods, the CV values for the frequent buyer levels are less than half of those for the idle buyer levels. 
However, the CV of the proposed method is consistently lower than that of the zero imputation method across all user segments. The decrease in the CV indicates an improvement in sensitivity for the outcome PR. This
improvement in sensitivity is largely attributable to the change in mean values.

\begin{figure}[]
\caption{Comparison of CV across user segments between the proposed imputation method and the zero imputation method for the treatment group. The tick values in the vertical axis are omitted for the restriction of disclosure.}
{\centering \includegraphics[width=1\linewidth]{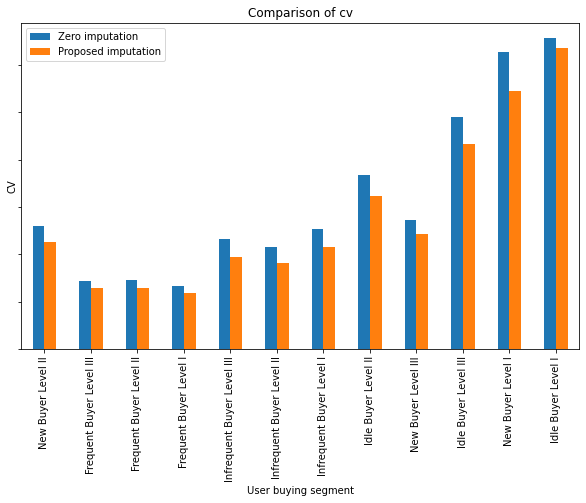}
}
\end{figure}

\hypertarget{discussion}{%
\section{Discussion}\label{discussion}}

Metrics provide strong evidence to support hypotheses in online
experimentation and hence reduce debates in the decision-making process.
This paper introduces the concept of dropout buyers and classifies users with incomplete metric values into two categories: visitors and dropout buyers.
For the analysis of incomplete metrics, we propose a cluster-based k-nearest neighbors-based imputation method. The proposed imputation
method considers both the experiment-specific features and users'
activities along their shopping paths. 
The proposed method incorporates uncertainty among missing values in the
outcome metrics using the k-nearest neighbors method. 
To facilitate efficient imputation in large-scale data sets in online experimentation, the proposed method employs a combination of stratification and clustering.
The stratification approach divides the entire large-scale data set into small
subsets to improve computation efficiency in the clustering step.
The clustering approach identifies inherent structure patterns to improve the performance of the k-nearest neighbors method within each cluster.

It is worth to remarking that the kNN method used in this work considered the average of responses in nearest neighbors. 
The weighted average of nearest neighbors has been proposed to suggest that different data points in the neighbor contribute differently to the decision based on their distances from the target point \citep{hechenbichler2004weighted}. That is, nearby data points, which are closer to the target in the neighbors, have higher influence on the decision than distant data points.
Moreover, one would incorporate the network structure information into the kNN for the networked A/B testing \citep{zhang2022locally}. 
Another direction for future research is to study ratio metrics \citep{jin2022toward} related to purchases in the proposed imputation framework. 
On the other hand, the proposed imputation method aims to impute missing values for each user with missing outcomes.
It would be interesting to categorize users with missing outcomes into various hubs and investigate the imputation strategy for each hub of users altogether.

\bibliographystyle{chicago}
\bibliography{ref}

\end{document}